%% file: main.tex
\pdfoutput=1
\documentclass[conference]{IEEEtran} 
\usepackage[utf8]{inputenc}

\usepackage[numbers]{natbib}

\usepackage{amsmath}
\usepackage{tikz}
\usetikzlibrary{positioning,calc,matrix,shapes.multipart,backgrounds}

\usepackage{pgfplots}

\usepackage{sidecap}
\sidecaptionvpos{figure}{c}

\usepackage{float}
\usepackage{multirow}

\title{Vehicle Detection from 3D Lidar Using Fully Convolutional Network}
\author{
\IEEEauthorblockN{Bo Li, Tianlei Zhang and Tian Xia}
\IEEEauthorblockA{Baidu Research – Institute for Deep Learning}
\texttt{\{libo24, zhangtianlei, xiatian\}@baidu.com}}
\begin{document}

\maketitle

\begin{abstract}
Convolutional network techniques have recently achieved great success in vision based detection tasks. This paper introduces the recent development of our research on transplanting the fully convolutional network technique to the detection tasks on 3D range scan data. Specifically, the scenario is set as the vehicle detection task from the range data of Velodyne 64E lidar. We proposes to present the data in a 2D point map and use a single 2D end-to-end fully convolutional network to predict the objectness confidence and the bounding boxes simultaneously. By carefully design the bounding box encoding, it is able to predict full 3D bounding boxes even using a 2D convolutional network. Experiments on the KITTI dataset shows the state-of-the-art performance of the proposed method. 
\end{abstract}

\begin{figure*}
\centering
\begin{tabular}{cc}
\includegraphics[trim=160 10 40 10, clip, width=0.45\textwidth]{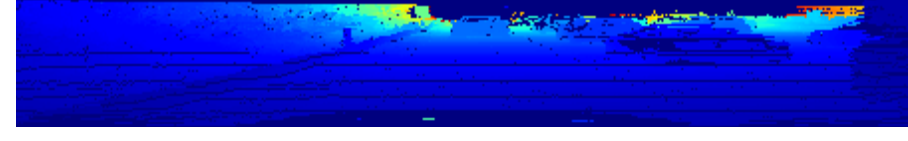} & \includegraphics[trim=160 10 40 10, clip, width=0.45\textwidth]{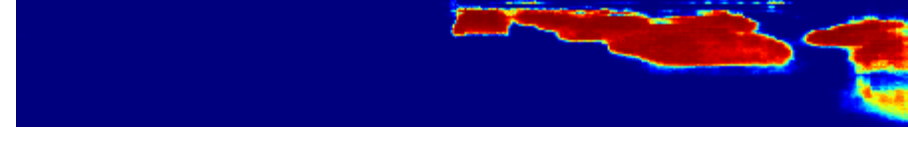} \\
(a) & (b) \\
\includegraphics[width=0.45\textwidth]{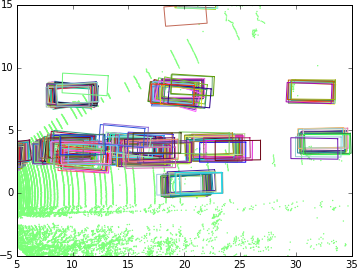} & \includegraphics[width=0.45\textwidth]{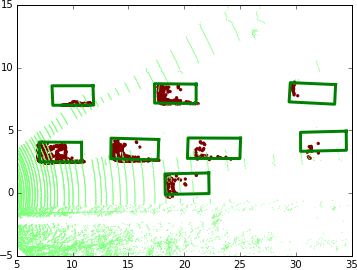} \\
(c) & (d)
\end{tabular}

\caption{Data visualization generated at different stages of the proposed approach. (a) The input point map, with the $d$ channel visualized. (b) The output confidence map of the objectness branch at $\mathbf{o}^a_\mathbf{p}$. Red denotes for higher confidence. (c) Bounding box candidates corresponding to all points predicted as positive, i.e. high confidence points in (b). (d) Remaining bounding boxes after non-max suppression. Red points are the groundtruth points on vehicles for reference. }
\label{fig:pipeline}
\end{figure*}

\section{Introduction}

For years of the development of robotics research, 3D lidars have been widely used on different kinds of robotic platforms.
Typical 3D lidar data present the environment information by 3D point cloud organized in a range scan. A large number of research have been done on exploiting the range scan data in robotic tasks including localization, mapping, object detection and scene parsing \cite{Levinson2010}.

In the task of object detection, range scans have an specific advantage over camera images in localizing the detected objects. Since range scans contain the spatial coordinates of the 3D point cloud by nature, it is easier to obtain the pose and shape of the detected objects. On a robotic system including both perception and control modules, e.g. an autonomous vehicle, accurately localizing the obstacle vehicles in the 3D coordinates is crucial for the subsequent planning and control stages. 

In this paper, we design a fully convolutional network (FCN) to detect and localize objects as 3D boxes from range scan data. FCN has achieved notable performance in computer vision based detection tasks. This paper transplants FCN to the detection task on 3D range scans. We strict our scenario as 3D vehicle detection for an autonomous driving system, using a Velodyne 64E lidar. The approach can be generalized to other object detection tasks on other similar lidar devices.

\section{Related Works}
\subsection{Object Detection from Range Scans}
Tranditional object detection algorithms propose candidates in the point cloud and then classify them as objects. A common category of the algorithms propose candidates by segmenting the point cloud into clusters. In some early works, rule-based segmentation is suggested for specific scene \cite{Himmelsbach2010, Moosmann2009, Douillard2011}. For example when processing the point cloud captured by an autonomous vehicle, simply removing the ground plane and cluster the remaining points can generate reasonable segmentation \cite{Himmelsbach2010, Douillard2011}. More delicate segmentation can be obtained by forming graphs on the point cloud \cite{Wang2012, Klasing2008, Papon2013, Triebel2006, Triebel2005}. The subsequent object detection is done by classifying each segments and thus is sometimes vulnerable to incorrect segmentation. To avoid this issue, \citet{Behley2013a} suggests to segment the scene hierarchically and keep segments of different scales. Other methods directly exhaust the range scan space to propose candidates to avoid incorrect segmentation. For example, \citet{Johnson1999} randomly samples points from the point cloud as correspondences. \citet{Wang} scan the whole space by a sliding window to generate proposals.

To classify the candidate data, some early researches assume known shape model and match the model to the range scan data \cite{Faugeras1986, Johnson1999}. In recent machine learning based detection works, a number of features have been hand-crafted to classify the candidates. \citet{Triebel2006, Wang2012, Teichman2011} use shape spin images, shape factors and shape distributions.  \citet{Teichman2011} also encodes the object moving track information for classification. \citet{Papon2013} uses FPFH. Other features include normal orientation, distribution histogram and etc. A comparison of features can be found in \cite{Behley2012}. Besides the hand-crafted features, \citet{Deuge2013, Lai2014} explore to  learn feature representation of point cloud via sparse coding. 

We would also like to mention that object detection on RGBD images \cite{Chen2015, Lin2013} is closely related to the topic of object detection on range scan. The depth channel can be interpreted as a range scan and naturally applies to some detection algorithms designed for range scan. On the other hand, numerous researches have been done on exploiting both depth and RGB information in object detection tasks. We omit detailed introduction about traditional literatures on RGBD data here but the proposed algorithm in this paper can also be generalized to RGBD data.

\subsection{Convolutional Neural Network on Object Detection}

The Convolutional Neural Network (CNN) has achieved notable succuess in the areas of object classification and detection on images. We mention some state-of-the-art CNN based detection framework here. R-CNN \cite{Girshick2014} proposes candidate regions and uses CNN to verify candidates as valid objects. OverFeat \cite{Sermanet2013}, DenseBox \cite{Huang2015} and YOLO \cite{Redmon2015} uses end-to-end unified FCN frameworks which predict the objectness confidence and the bounding boxes simultaneously over the whole image. Some research has also been focused on applying CNN on 3D data. For example on RGBD data, one common aspect is to treat the depthmaps as image channels and use 2D CNN for classification or detection \cite{Gupta2014, Schwarz2015, Socher2012}. For 3D range scan some works discretize point cloud along 3D grids and train 3D CNN structure for classification \cite{Wu2015, Maturana2015}. These classifiers can be integrated with region proposal method like sliding window \cite{Song2014} for detection tasks. The 3D CNN preserves more 3D spatial information from the data than 2D CNN while 2D CNN is computationally more efficient. 

In this paper, our approach project range scans as 2D maps similar to the depthmap of RGBD data. The frameworks of \citet{Huang2015, Sermanet2013} are transplanted to predict the objectness and the 3D object bounding boxes in a unified end-to-end manner.

\section{Approach}

\subsection{Data Preparation}
We consider the point cloud captured by the Velodyne 64E lidar. Like other range scan data, points from a Velodyne scan can be roughly projected and discretized into a 2D point map, using the following projection function.
\begin{equation}
\begin{aligned}
    \theta &= \textrm{atan2} (y, x)\\
    \phi &= \arcsin (z / \sqrt{x^2 + y^2 + z^2})\\
    r &= \lfloor \theta / \Delta \theta \rfloor \\
    c &= \lfloor \phi / \Delta \phi \rfloor
\end{aligned}
\label{eq:projection}
\end{equation}
where $\mathbf{p} = (x, y, z)^\top$ denotes a 3D point and $(r, c)$ denotes the 2D map position of its projection. $\theta$ and $\phi$ denote the azimuth and elevation angle when observing the point. $\Delta \theta$ and $\Delta \phi$ is the average horizontal and vertical angle resolution between consecutive beam emitters, respectively. The projected point map is analogous to cylindral images. We fill the element at $(r, c)$ in the 2D point map with 2-channel data $(d, z)$ where $d = \sqrt{x^2 + y^2}$. Note that $x$ and $y$ are coupled as $d$ for rotation invariance around $z$. An example of the $d$ channel of the 2D point map is shown in Figure \ref{fig:pipeline}a. Rarely some points might be projected into a same 2D position, in which case the point nearer to the observer is kept. Elements in 2D positions where no 3D points are projected into are filled with $(d, z) = (0, 0)$. 

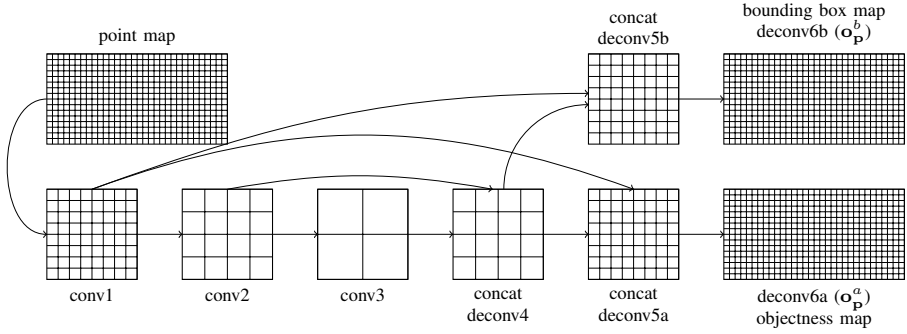
\begin{SCfigure*}
\centering
\input{net.tikz}
\caption{The proposed FCN structure to predict vehicle objectness and bounding box simultaneously. The output feature map of conv1/deconv5a, conv1/deconv5b and conv2/deconv4 are first concatenated and then ported to their consecutive layers, respectively.}
\label{fig:fcn_structure}
\end{SCfigure*}

\subsection{Network Architecture}
\label{sec:network_architecture}

The trunk part of the proposed CNN architecture is similar to \citet{Huang2015, Long}. As illustrated in Figure \ref{fig:fcn_structure}, the CNN feature map is down-sampled consecutively in the first 3 convolutional layers and up-sampled consecutively in deconvolutional layers. Then the trunk splits at the 4th layer into a objectness classification branch and a 3D bounding box regression branch. We describe its implementation details as follows:
\begin{itemize}
\item The input point map, output objectness map and bounding box map are of the same width and height, to provide point-wise prediction. Each element of the objectness map predicts whether its corresponding point is on a vehicle. If the corresponding point is on a vehicle, its corresponding element in the bounding box map predicts the 3D bounding box of the belonging vehicle. Section \ref{prediction_encoding} explains how the objectness and bounding box is encoded.
\item In conv1, the point map is down-sampled by 4 horizontally and 2 vertically. This is because for a point map captured by Velodyne 64E, we have approximately $\Delta \phi = 2 \Delta \theta$, i.e. points are denser on horizotal direction. Similarly, the feature map is up-sampled by this factor of (4, 2) in deconv6a and deconv6b, respectively. The rest conv/deconv layers all have equal horizontal and vertical resolution, respectively, and use squared strides of $(2, 2)$ when up-sampling or down-sampling.
\item The output feature map pairs of conv3/deconv4, conv2/deconv5a, conv2/deconv5b are of the same sizes, respectively. We concatenate these output feature map pairs before passing them to the subsequent layers. This follows the idea of \citet{Long}. Combining features from lower layers and higher layers improves the prediction of small objects and object edges. 
\end{itemize}

\subsection{Prediction Encoding}
\label{prediction_encoding}
\begin{SCfigure*}
\centering
\begin{tabular}{cc}
\input{box-conversion.tikz} & \input{rotation-invariance.tikz} \\
(a) & (b)
\end{tabular}
\caption{(a) Illustration of (\ref{eq:transform}). For each vehicle point $\mathbf{p}$, we define a specific coordinate system which is centered at $\mathbf{p}$. The $x$ axis ($\mathbf{r}_x$) of the coordinate system is along with the ray from Velodyne origin to $\mathbf{p}$ (dashed line). (b) An example illustration about the rotation invariance when observing a vehicle. Vehicle A and B have same appearance. See (\ref{eq:transform}) in Section \ref{prediction_encoding} for details. }
\label{fig:transform}
\end{SCfigure*}

We now describe how the output feature maps are defined. The objectness map deconv6a consists of 2 channels corresponding to foreground, i.e. the point is on a vehicle, and background. The 2 channels are normalized by softmax to denote the confidence.

The encoding of the bounding box map requires some extra conversion. Consider a lidar point $\mathbf{p} = (x, y, z)$ on a vehicle. Its observation angle is $(\theta, \phi)$ by (\ref{eq:projection}). We first denote a rotation matrix $\mathbf{R}$ as
\begin{equation}
\mathbf{R} = \mathbf{R}_z(\theta) \mathbf{R}_y(\phi) 
\end{equation}
where $\mathbf{R}_z(\theta)$ and $\mathbf{R}_y(\phi)$ denotes rotations around $z$ and $y$ axes respectively. If denote $\mathbf{R}$ as $(\mathbf{r}_x, \mathbf{r}_y, \mathbf{r}_z)$, $\mathbf{r}_x$ is of the same direction as $\mathbf{p}$ and $\mathbf{r}_y$ is parallel with the horizontal plane. Figure \ref{fig:transform}a illustrate an example on how $\mathbf{R}$ is formed. A bounding box corner $\mathbf{c}_\mathbf{p} = (x_c, y_c, z_c)$ is thus transformed as:
\begin{equation}
\mathbf{c}_\mathbf{p}' = \mathbf{R}^\top ( \mathbf{c}_\mathbf{p} - \mathbf{p} )
\label{eq:transform}
\end{equation}
Our proposed approach uses $\mathbf{c}_\mathbf{p}'$ to encode the bounding box corner of the vehicle which $\mathbf{p}$ belongs to. The full bounding box is thus encoded by concatenating 8 corners in a 24d vector as 
\begin{equation}
\mathbf{b}_\mathbf{p}' = (\mathbf{c}_{\mathbf{p}, 1}'^\top, \mathbf{c}_{\mathbf{p}, 2}'^\top, \dots, \mathbf{c}_{\mathbf{p}, 8}'^\top)^\top
\label{eq:box_encoding}
\end{equation}
Corresponding to this 24d vector, deconv6b outputs a 24-channel feature map accordingly.

The transform (\ref{eq:transform}) is designed due to the following two reasons:
\begin{itemize}
\item \textit{Translation part} Compared to $\mathbf{c}_\mathbf{p}$ which distributes over the whole lidar perception range, e.g. $[-100\textrm{m}, 100\textrm{m}]\times[-100\textrm{m}, 100\textrm{m}]$ for Velodyne, the corner offset $\mathbf{c}_\mathbf{p} - \mathbf{p}$ distributes in a much smaller range, e.g. within size of a vehicle. Experiments show that it is easier for the CNN to learn the latter case.
\item \textit{Rotation part} $\mathbf{R}^\top$ ensures the rotation invariance of the corner coordinate encoding. When a vehicle is moving around a circle and one observes it from the center, the appearance of the vehicle does not change in the observed range scan but the bounding box coordinates vary in the range scan coordinate system. Since we would like to ensure that same appearances result in same bounding box prediction encoding, the bounding box coordinates are rotated by $\mathbf{R}^\top$ to be invariant. Figure \ref{fig:transform}b illustrates a simple case. Vehicle A and B have the same appearance for an observer at the center, i.e. the right side is observed. Vehicle C has a difference appearance, i.e. the rear-right part is observed. With the conversion of (\ref{eq:transform}), the bounding box encoding $\mathbf{b}_\mathbf{p}'$ of A and B are the same but that of C is different.
\end{itemize}

\subsection{Training Phase}
\subsubsection{Data Augmentation}
Similar to the training phase of a CNN for images, data augmentation significantly enhances the network performance. For the case of images, training data are usually augmented by randomly zooming or rotating the original images to synthesis more training samples. For the case of range scans, simply applying these operations results in variable $\Delta \theta$ and $\Delta \phi$ in (\ref{eq:projection}), which violates the geometry property of the lidar device. To synthesis geometrically correct 3D range scans, we randomly generate a 3D transform near identity. Before projecting point cloud by (\ref{eq:projection}), the random transform is applied the point cloud. The translation component of the transform results in zooming effect of the synthesized range scan. The rotation component results in rotation effect of the range scan. 

\subsubsection{Multi-Task Training}
As illustrated Section \ref{sec:network_architecture}, the proposed network consists of one objectness classification branch and one bounding box regression branch. We respectively denote the losses of the two branches in the training phase. As notation, denote $\mathbf{o}^a_\mathbf{p}$ and $\mathbf{o}^b_\mathbf{p}$ as the feature map output of deconv6a and deconv6b corresponding to point $\mathbf{p}$ respectively. Also denote $\mathcal{P}$ as the point cloud and $\mathcal{V} \subset \mathcal{P}$ as all points on all vehicles.

The loss of the objectness classification branch corresponding to a point $\mathbf{p}$ is denoted as a softmax loss
\begin{equation}
\begin{aligned}
    \mathcal{L}_\textrm{obj}(p) &= - \log (p_\mathbf{p}) \\
    p_\mathbf{p} &= \cfrac{\exp (-\mathbf{o}^a_{\mathbf{p}, l_\mathbf{p}})}{ \sum_{l \in \{0, 1\}}{\exp (-\mathbf{o}^a_{\mathbf{p}, l})}}
\end{aligned}
\label{eq:object_loss}
\end{equation}
where $l_\mathbf{p} \in \{0, 1\}$ denotes the groundtruth objectness label of $\mathbf{p}$, i.e. 0 as background and 1 as a point on vechicles.  $o^a_{\mathbf{p}, \star}$ denotes the deconv6a feature map output of channel $\star$ for point $\mathbf{p}$.

The loss of the bounding box regression branch corresponding to a point $\mathbf{p}$ is denoted as a L2-norm loss
\begin{equation}
\begin{aligned}
    \mathcal{L}_\textrm{box}(p) = \| \mathbf{o}^b_{\mathbf{p}} - \mathbf{b}_\mathbf{p}' \|^2
\end{aligned}
\label{eq:box_loss}
\end{equation}
where $\mathbf{b}_\mathbf{p}'$ is a 24d vector denoted in (\ref{eq:box_encoding}). Note that $\mathcal{L}_\textrm{box}$ is only computed for those points on vehicles. For non-vehicle points, the bounding box loss is omitted. 

\subsubsection{Training strategies}
Compared to positive points on vehicles, negative (background) points account for the majority portion of the point cloud. Thus if simply pass all objectness losses in (\ref{eq:object_loss}) in the backward procedure, the network prediction will significantly bias towards negative samples. To avoid this effect, losses of positive and negative points need to be balanced. Similar balance strategies can be found in \citet{Huang2015} by randomly discarding redundant negative losses. In our training procedure, the balance is done by keeping all negative losses but re-weighting them using
\begin{equation}
w_1(\mathbf{p}) = \left \{
\begin{aligned}
    & k |\mathcal{V}| / (|\mathcal{P}| - |\mathcal{V}|) & & {\mathbf{p} \in \mathcal{P} - \mathcal{V}} \\
    & 1 & & {\mathbf{p} \in \mathcal{V}}
\end{aligned}
\right.
\end{equation}
which denotes that the re-weighted negative losses are averagely equivalent to losses of $k |\mathcal{V}|$ negative samples. In our case we choose $k = 4$. Compared to randomly discarding samples, the proposed balance strategy keeps more information of negative samples.

Additionally, near vehicles usually account for larger portion of points than far vehicles and occluded vehicles. Thus vehicle samples at different distances also need to be balanced. This helps avoid the prediction to bias towards near vehicles and neglect far vehicles or occluded vehicles. Denote $n(\mathbf{p})$ as the number of points belonging to the same vehicle with $\mathbf{p}$. Since the 3D range scan points are almost uniquely projected onto the point map. $n(\mathbf{p})$ is also the area of the vehicle of $\mathbf{p}$ on the point map. Denote $\bar{n}$ as the average number of points of vehicles in the whole dataset. We re-weight $\mathcal{L}_\textrm{obj}(\mathbf{p})$ and $\mathcal{L}_\textrm{box}(\mathbf{p})$ by $w_2$ as
\begin{equation}
w_2(\mathbf{p}) = \left \{
\begin{aligned}
    &\bar{n} / n(\mathbf{p}) & & {\mathbf{p} \in \mathcal{V}} \\
    &1 & & {\mathbf{p} \in \mathcal{P} - \mathcal{V}}
\end{aligned}
\right.
\end{equation}

Using the losses and weights designed above, we accumulate losses over deconv6a and deconv6b for the final training loss
\begin{equation}
    \mathcal{L} = \sum_{\mathbf{p} \in \mathcal{P}} w_1(\mathbf{p}) w_2(\mathbf{p}) \mathcal{L}_\textrm{obj}(\mathbf{p}) + w_\textrm{box} \sum_{\mathbf{p} \in \mathcal{V}} w_2(\mathbf{p}) \mathcal{L}_\textrm{box}(\mathbf{p})
\end{equation}
with $w_\textrm{box}$ used to balance the objectness loss and the bounding box loss.

\subsection{Testing Phase}
During the test phase, a range scan data is fed to the network to produce the objectness map and the bounding box map. For each point which is predicted as positive in the objectness map, the corresponding output $\mathbf{o}^b_{\mathbf{p}}$ of the bounding box map is splitted as $\mathbf{c}_{\mathbf{p}, i}', i = 1, \dots, 8$. $\mathbf{c}_{\mathbf{p}, i}'$ is then converted to box corner $\mathbf{c}_{\mathbf{p}, i}$ by the inverse transform of (\ref{eq:transform}). We denote each bounding box candidates as a 24d vector $\mathbf{b}_\mathbf{p} = (\mathbf{c}_{\mathbf{p}, 1}^\top, \mathbf{c}_{\mathbf{p}, 2}^\top, \cdots, \mathbf{c}_{\mathbf{p}, 8}^\top)^\top$. The set of all bounding box candidates is denoted as $\mathbf{B} = \{\mathbf{b}_\mathbf{p} | \mathbf{o}^a_{\mathbf{p}, 1} > \mathbf{o}^a_{\mathbf{p}, 0} \}$. Figure \ref{fig:pipeline}c shows the bounding box candidates of all the points predicted as positive. 

We next cluster the bounding boxes and prune outliers by a non-max suppression strategy. Each bounding box $\mathbf{b}_\mathbf{p}$ is scored by counting its neighbor bounding boxes in $\mathbf{B}$ within a distance $\delta$, denoted as $\#\{ \mathbf{x} \in \mathbf{B} | \| \mathbf{x} - \mathbf{b}_\mathbf{p} \| < \delta \}$. Bounding boxes are picked from high score to low score. After one box is picked, we find out all points inside the bounding box and remove their corresponding bounding box candidates from $\mathbf{B}$. Bounding box candidates whose score is lower than 5 is discarded as outliers. Figure \ref{fig:pipeline}d shows the picked bounding boxes for Figure \ref{fig:pipeline}a.

\begin{figure*}
\centering
\begin{tabular}{cc}
\includegraphics[width=0.45\textwidth]{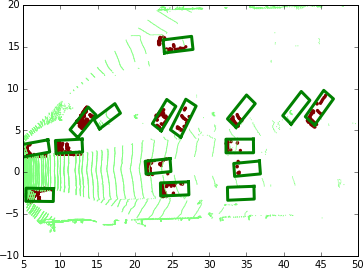} & \includegraphics[width=0.45\textwidth]{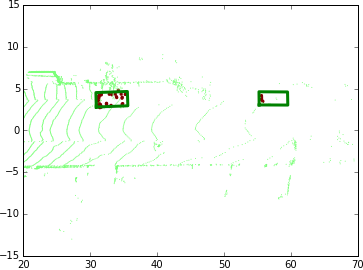} \\
(a) & (b)
\end{tabular}
\caption{More examples of the detection results. See Section \ref{sec:offline} for details. (a) Detection result on a congested traffic scene. (b) Detection result on far vehicles. }
\label{fig:samples}
\end{figure*}

\section{Experiments}
Our proposed approach is evaluated on the vehicle detection task of the KITTI object detection benchmark \cite{Geiger2012}. This benchmark originally aims to evaluate object detection of vehicles, pedestrians and cyclists from images. It contains not only image data but also corresponding Velodyne 64E range scan data. The groundtruth labels include both 2D object bounding boxes on images and its corresponding 3D bounding boxes, which provides sufficient information to train and test detection algorithm on range scans. The KITTI training dataset contains 7500+ frames of data. We randomly select 6000 frames in our experiments to train the network and use the rest 1500 frames for detailed offline validation and analysis. The KITTI online evaluation is also used to compare the proposed approach with previous related works.

For simplicity of the experiments, we focus our experiemts only on the \textit{Car} category of the data. In the training phase, we first label all 3D points inside any of the groundtruth car 3D bounding boxes as foreground vehicle points. Points from objects of categories like \textit{Truck} or \textit{Van} are labeled to be ignored from $\mathcal{P}$ since they might confuse the training. The rest of the points are labeled as background. This forms the label $l_\mathbf{p}$ in (\ref{eq:object_loss}). For each foreground point, its belonging bounding box is encoded by (\ref{eq:box_encoding}) to form the label $\mathbf{b}_\mathbf{p}'$ in (\ref{eq:box_loss}).

The experiments are based on the Caffe \cite{Jia2014} framework. In the KITTI object detection benchmark, images are captured from the front camera and range scans percept a $360^{\circ}$ FoV of the environment. The benchmark groundtruth are only provided for vehicles inside the image. Thus in our experiment we only use the front part of a range scan which overlaps with the FoV of the front camera.

The KITTI benchmark divides object samples into three difficulty levels according to the size and the occlusion of the 2D bounding boxes in the image space. A detection is accepted if its image space 2D bounding box has at least 70\% overlap with the groundtruth. Since the proposed approach naturally predicts the 3D bounding boxes of the vehicles, we evaluate the approach in both the image space and the world space in the offline validation. Compared to the image space, metric in the world space is more crucial in the scenario of autonomous driving. Because for example many navigation and planning algorithms take the bounding box in world space as input for obstacle avoidance. Section \ref{sec:offline} describes the evaluation in both image space and world space in our offline validation. In Section \ref{sec:online}, we compare the proposed approach with several previous range scan detection algorithms via the KITTI online evaluation system.

\begin{table}
    \centering
    \caption{Performance in Average Precision and Average Orientation Similarity for the Offline Evaluation}
    \begin{tabular}{cccc}
    \hline\hline
     & Easy & Moderate & Hard \\
    \hline
    Image Space (AP) & 74.1\% & 71.0\% & 70.0\% \\
    Image Space (AOS) & 73.9\% & 70.9\% & 69.9\%\\
    World Space (AP) & 77.3\% & 72.4\% & 69.4\% \\
    World Space (AOS) & 77.2\% & 72.3\% & 69.4\% \\
    \hline
    \end{tabular}
    \label{tab:offline}
\end{table}

\begin{figure}
    \centering
    \includegraphics[width=0.38\textwidth]{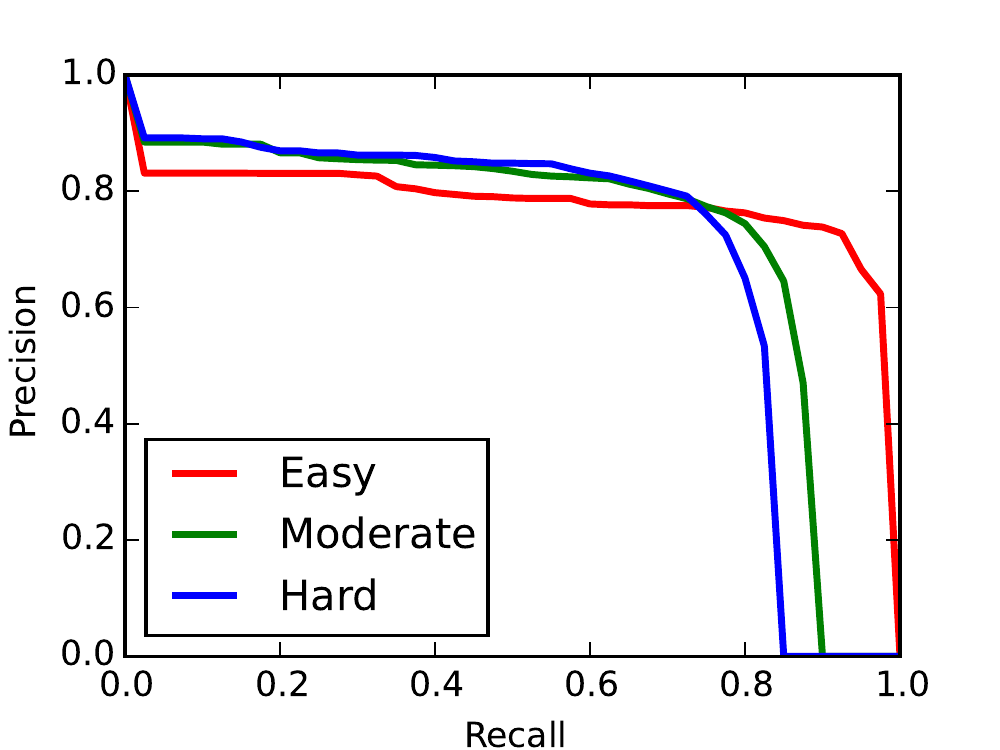}
    \caption{Precision-recall curve in the offline evaluation, measured by the world space criterion. See Section \ref{sec:offline}. }
    \label{fig:curve}
\end{figure}

\subsection{Performane Analysis on Offline Evaluation}
\label{sec:offline}
We analyze the detection performance on our custom offline evaluation data selected from the KITTI training dataset, whose groundtruth labels are accessable to public. To obtain an equivalent 2D bounding box for the original KITTI criterion in the image space, we projected the 3D bounding box into the image space and take the minimum 2D bounding rectangle as the 2D bounding box. For the world space evaluation, we project the detected and the groundtruth 3D bounding boxes onto the ground plane and compute their overlap. The world space criterion also requires at least 70\% overlap to accept a detection. The performance of the approach is measured by the Average Precision (AP) and the Average Orientation Similarity (AOS) \cite{Geiger2012}. The AOS is designed to jointly measure the precision of detection and orientation estimation.

Table \ref{tab:offline} lists the performance evaluation. Note that the world space criterion results in slightly better performance than the image space criterion. This is because the user labeled 2D bounding box trends to be tighter than the 2D projection of the 3D bounding boxes in the image space, especially for vehicles observed from their diagonal directions. This size difference diminishes the overlap between the detection and the groundtruth in the image space.

Like most detection approaches, there is a noticeable drop of performance from the easy evaluation to the moderate and hard evaluation. The minimal pixel height for easy samples is 40px. This approximately corresponds to vehicles within 28m. The minimal height for moderate and hard samples is 25px, corresponding to minimal distance of 47m. As shown in Figure \ref{fig:samples} and Figure \ref{fig:pipeline}, some vehicles farther than 40m are scanned by very few points and are even difficult to recognize for human. This results in the performance drop for moderate and hard evalutaion.

Figure \ref{fig:curve} shows the precision-recall curve of the world space criterion as an example. Precision-recall curves of the other criterion are similar and omitted here. Figure \ref{fig:samples}a shows the detection result on a congested traffic scene with more than 10 vehicles in front of the lidar.  Figure \ref{fig:samples}b shows the detection result cars farther than 50m. Note that our algorithm predicts the completed bounding box even for vehicles which are only partly visible. This significantly differs from previous proposal-based methods and can contribute to stabler object tracking and path planning results. For the easy evaluation, the algorithm detects almost all vehicles, even occluded. This is also illustrated in Figure \ref{fig:curve} where the maximum recall rate is higher than 95\%. The approach produces false-positive detection in some occluded scenes, which is illustrated in Figure \ref{fig:samples}a for example.

\subsection{Related Work Comparison on the Online Evaluation}
\label{sec:online}
There have been several previous works in range scan based detection evaluated on the KITTI platform. Readers might find that the performance of these works ranks much lower compared to the state-of-the-art vision-based approaches. We explain this by two reasons. First, the image data have much higher resolution which significantly enhance the detection performance for far and occluded objects. Second, the image space based criterion does not reflect the advantage of range scan methods in localizing objects in full 3D world space.  Related explanation can also be found from \citet{Wang}. Thus in this experiments, we only compare the proposed approach with range scan methods of \citet{Wang, Behley2013a, thesisSTUDIENARBEITPlotkin}. These three methods all use traditional features for classification. \citet{Wang} performs a sliding window based strategy to generate candidates and \citet{Behley2013a, thesisSTUDIENARBEITPlotkin} segment the point cloud to generate detection candidates.

Table \ref{tab:online} shows the performance of the methods in AP and AOS reported on the KITTI online evaluation. The detection AP of our approach outperforms the other methods in the easy task, which well illustrates the advantage of CNN in representing rich features on near vehicles. In the moderate and hard detection tasks, our approach performs with similar AP as \citet{Wang}. Because vehicles in these tasks consist of too few points for CNN to embed complicated features. For the joint detection and orientation estimation evaluation, only our approach and CSoR support orientation estimation and our approach significantly wins the comparison in AOS.

\begin{table}
    \centering
    \caption{Performance Comparison in Average Precision and Average Orientation Similarity for the Online Evaluation}
    \begin{tabular}{ccccc}
    \hline\hline
    & & Easy & Moderate & Hard \\
    \hline
    \multirow{4}{*}{Image Space (AP)} & Proposed & 60.3\% & 47.5\% & 42.7\% \\
    & Vote3D & 56.8\% & 48.0\% & 42.6\%\\
    & CSoR & 34.8\% & 26.1\% & 22.7\% \\
    & mBoW & 36.0\% & 23.8\% & 18.4\% \\
    \hline
    \multirow{2}{*}{Image Space (AOS)} & Proposed & 59.1\% & 45.9\% & 41.1\% \\
    & CSoR & 34.0\% & 25.4\% & 22.0\% \\
    \hline

    \end{tabular}
    \label{tab:online}
\end{table}

\section{Conclusions}
Although attempts have been made in a few previous research to apply deep learning techniques on sensor data other than images, there is still a gap inbetween this state-of-the-art computer vision techniques and the robotic perception research. To the best of our knowledge, the proposed approach is the first to introduce the FCN detection techniques into the perception on range scan data, which results in a neat and end-to-end detection framework. In this paper we only evaluate the approach on 3D range scan from Velodyne 64E but the approach can also be applied on 3D range scan from similar devices. By accumulating more training data and design deeper network, the detection performance can be even further improved. 

\section{Acknowledgement}
The author would like to acknowledge the help from Ji Liang, Lichao Huang, Degang Yang, Haoqi Fan and Yifeng Pan in the research of deep learning. Thanks also go to Ji Tao, Kai Ni and Yuanqing Lin for their support.

\bibliographystyle{plainnat}
\bibliography{main}
\end{document}

%% file: net.tikz
\begin{tikzpicture}[scale=0.6]

\draw[ultra thin] (3,3) grid[step=0.125] (7, 5); 
\draw[thin] (3,3) rectangle (7, 5) node [above, font=\scriptsize] at (5, 5) {point map};
\draw[ultra thin]  (3,0) grid[step=0.25] (5, 2);
\draw[thin] (3,0) rectangle (5, 2) node [below, font=\scriptsize] at (4, 0) {conv1};
\draw[ultra thin]  (6,0) grid[step=0.5] (8, 2);
\draw[thin] (6,0) rectangle (8, 2) node [below, font=\scriptsize] at (7, 0) {conv2};
\draw[ultra thin]  (9,0) grid[step=1] (11, 2);
\draw[thin] (9,0) rectangle (11, 2) node [below, font=\scriptsize] at (10, 0) {conv3};
\draw[ultra thin]  (12,0) grid[step=0.5] (14, 2);
\draw[thin] (12,0) rectangle (14, 2) node [below, font=\scriptsize,  align=center] at (13, 0) {concat \\ deconv4};
\draw[ultra thin]  (15,0) grid[step=0.25] (17, 2);
\draw[thin] (15,0) rectangle (17, 2) node [below, font=\scriptsize,  align=center] at (16, 0) {concat \\ deconv5a};
\draw[ultra thin] (18,0) grid[step=0.125] (22, 2);
\draw[thin] (18,0) rectangle (22, 2) node [below, font=\scriptsize, align=center] at (20, 0) {deconv6a ($\mathbf{o}^a_\mathbf{p}$)\\ objectness map};
\draw[ultra thin]  (15,3) grid[step=0.25] (17, 5);
\draw[thin] (15,3) rectangle (17, 5) node [above, font=\scriptsize,  align=center] at (16, 5) {concat \\ deconv5b};
\draw[ultra thin] (18,3) grid[step=0.125] (22, 5);
\draw[thin] (18,3) rectangle (22, 5) node [above, align=center, font=\scriptsize] at (20, 5) {bounding box map\\ deconv6b ($\mathbf{o}^b_\mathbf{p}$)};

\draw[ultra thin,->] (3,4) to[out=180,in=180] (3,1);
\draw[ultra thin,->] (4,2) to[out=20,in=160] (16,2);
\draw[ultra thin,->] (7,2) to[out=10,in=170] (12.875,2);
\draw[ultra thin,->] (4,2) to[out=20,in=180] (15,4.125);
\draw[ultra thin,->] (5,1) to (6,1);
\draw[ultra thin,->] (8,1) to (9,1);
\draw[ultra thin,->] (11,1) to (12,1);
\draw[ultra thin,->] (14,1) to (15,1);
\draw[ultra thin,->] (17,1) to (18,1);
\draw[ultra thin,->] (17,4) to (18,4);
\draw[ultra thin,->] (13.125,2) to[out=90,in=180] (15,3.875);
\end{tikzpicture}

%% file: box-conversion.tikz
\begin{tikzpicture}[scale=1.5]


\begin{axis}[view={-50}{-10}, axis equal, hide axis,
    ]
\addplot3 coordinates {(0, 0, 0)};
\node at (axis cs:0, 0, 0) { \includegraphics[width=2cm] {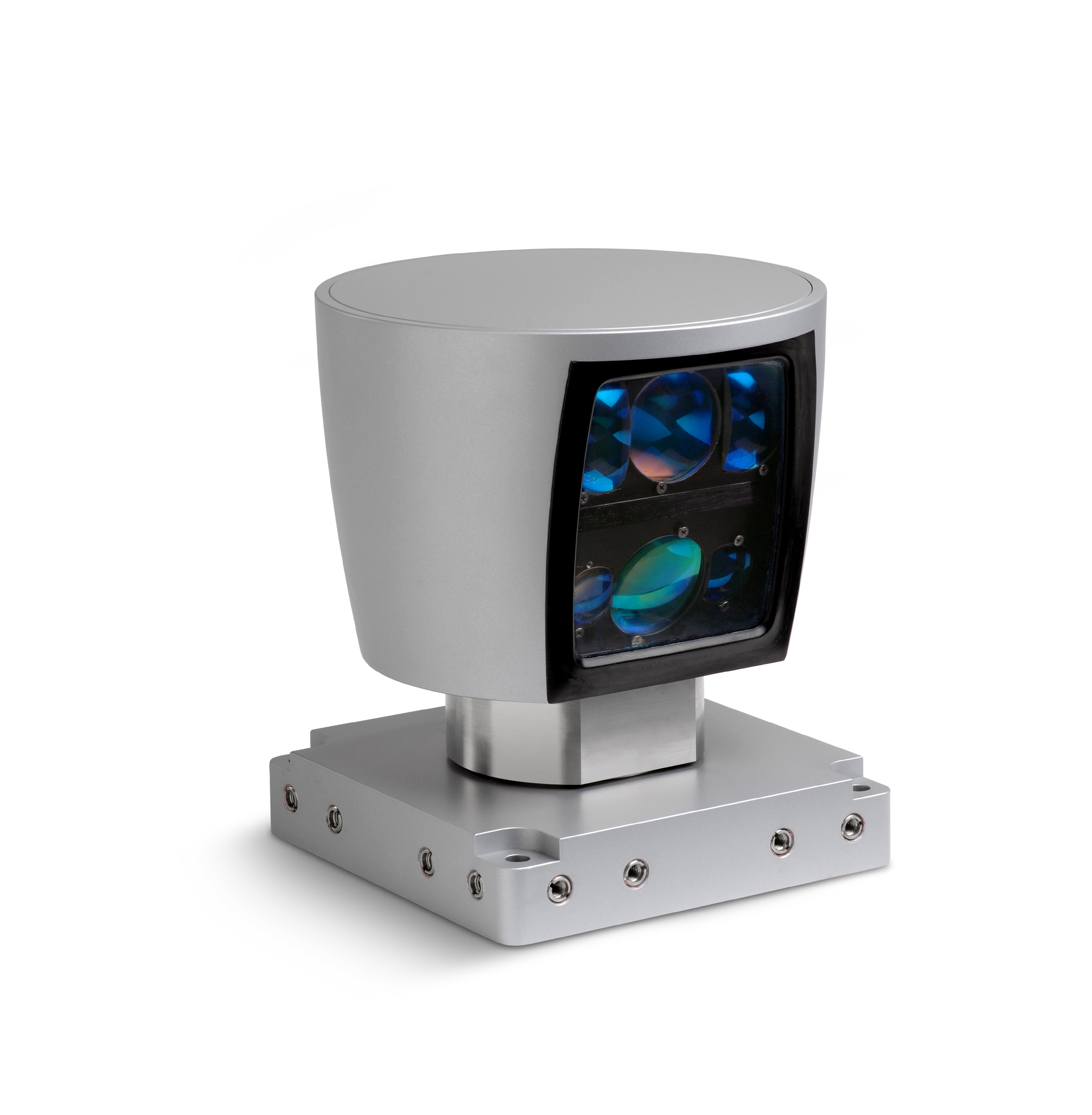}};

\draw[->, thin](axis cs:0, 0, 0)--(axis cs:3, 0, 0);
\draw[->, thin](axis cs:0, 0, 0)--(axis cs:0, 3, 0);
\draw[->, thin](axis cs:0, 0, 0)--(axis cs:0, 0, 3);
\addplot3[only marks, dotted, mark=*, mark size=1, gray] table {fig-vehicle-points.txt};
\addplot3 coordinates {(12, -1.03, -1.51)} node [below, font=\scriptsize] {$\mathbf{p}$};
\draw[dashed, ultra thin, gray](axis cs:0, 0, 0)--(axis cs:12, -1.03, -1.51);
\draw[->, thin](axis cs: 12, -1.03, -1.51)--(axis cs:  14.96579203,  -1.28456382,  -1.8831955 ) node [right, font=\scriptsize]{$\mathbf{r}_x$};
\draw[->, thin](axis cs: 12, -1.03, -1.51)--(axis cs:   11.74344334,  -4.01900965,  -1.51   ) node [right, font=\scriptsize]{$\mathbf{r}_y$};
\draw[->, thin](axis cs: 12, -1.03, -1.51)--(axis cs: 12.37182831,  -1.06191526,   1.46669702) node [above, font=\scriptsize]{$\mathbf{r}_z$};

\draw [gray, ultra thin](axis cs: 11.550143802035269, -1.536787856520008, -0.23645726895633756)--(axis cs:11.609739405055944, -0.10807268571676842, -0.23457496518169185)--(axis cs:15.455827754289047, -0.2686188267111633, -0.17775955837105695)--(axis cs: 15.39623215126837, -1.6973339975144028, -0.17964186214570266)--(axis cs: 11.550143802035269, -1.536787856520008, -0.23645726895633756);
\draw [gray, ultra thin](axis cs: 11.568932664240542, -1.5358850310868548, -1.5063055466456652)--(axis cs:11.628528267261219, -0.1071698602836153, -1.5044232428710194)--(axis cs:15.47461661649432, -0.2677160012780102, -1.4476078360603846)--(axis cs:15.415021013473645, -1.6964311720812497, -1.4494901398350302)--(axis cs: 11.568932664240542, -1.5358850310868548, -1.5063055466456652);
\draw [gray, ultra thin](axis cs: 11.550143802035269, -1.536787856520008, -0.23645726895633756)--(axis cs:11.609739405055944, -0.10807268571676842, -0.23457496518169185)--(axis cs:11.628528267261219, -0.1071698602836153, -1.5044232428710194)--(axis cs: 11.568932664240542, -1.5358850310868548, -1.5063055466456652)--(axis cs: 11.550143802035269, -1.536787856520008, -0.23645726895633756);
\draw [gray, ultra thin](axis cs:15.455827754289047, -0.2686188267111633, -0.17775955837105695)--(axis cs: 15.39623215126837, -1.6973339975144028, -0.17964186214570266)--(axis cs:15.415021013473645, -1.6964311720812497, -1.4494901398350302)--(axis cs:15.47461661649432, -0.2677160012780102, -1.4476078360603846)--(axis cs:15.455827754289047, -0.2686188267111633, -0.17775955837105695);

\end{axis}
\pgfresetboundingbox
\path (0.5, 2) rectangle (5.8, 4.5);

\end{tikzpicture}

%% file: rotation-invariance.tikz
\begin{tikzpicture}[scale=0.7]
\node [rotate=0] at (0, 2) { \includegraphics[width=1.5cm] {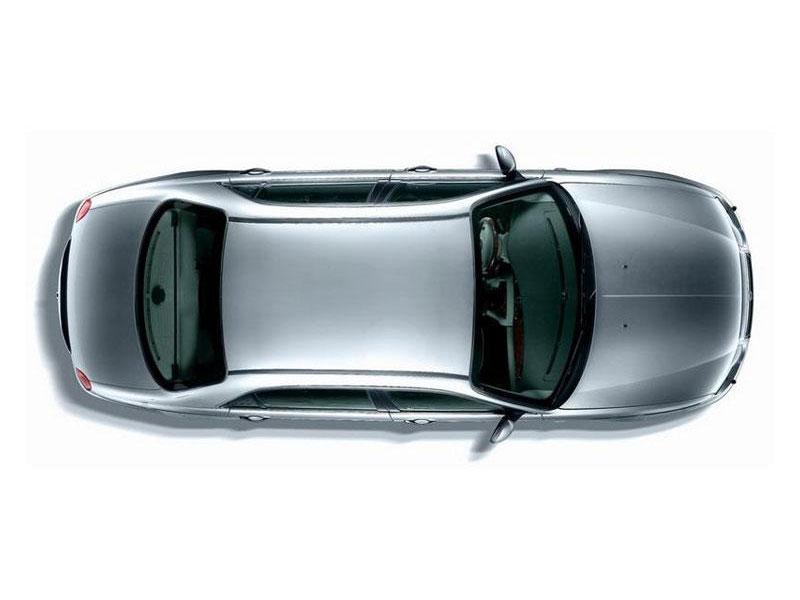}};
\node [font=\footnotesize] at (0, 3) {A};

\node [rotate=0] at (2.2, 2) { \includegraphics[width=1.5cm] {fig-vehicle.jpg}};
\node [font=\footnotesize] at (2, 3) {C};

\node [rotate=60] at (-1.732, 1) { \includegraphics[width=1.5cm] {fig-vehicle.jpg}};
\node [font=\footnotesize] at (-2.5, 1.5) {B};

\filldraw (0,0) circle (2pt);
\draw [->] (0, 0)--(1, 0);
\draw [->] (0, 0)--(0, 1);
\end{tikzpicture}